\documentclass{article}

\PassOptionsToPackage{numbers,sort&compress,square}{natbib}
 \usepackage[final]{neurips_2025}

\usepackage[utf8]{inputenc} 
\usepackage[T1]{fontenc}    
\usepackage{hyperref}       
\usepackage{url}            
\usepackage{booktabs}       
\usepackage{amsfonts}       
\usepackage{nicefrac}       
\usepackage{microtype}      
\usepackage{xcolor}         
\usepackage{graphicx}
\usepackage{array}
\usepackage{colortbl} 
\usepackage{adjustbox}
\usepackage{listings}
\usepackage{caption}

\usepackage{float} 
\usepackage{placeins}  
\usepackage{dblfloatfix} 
\usepackage{amsmath}
\usepackage{amssymb}
\usepackage{mathtools}
\usepackage{amsthm}
\usepackage{subcaption}
\usepackage{enumitem}
\usepackage{multirow}

\title{MGA-VQA: Secure and Interpretable Graph-Augmented Visual Question Answering with Memory-Guided Protection Against Unauthorized Knowledge Use}

\author{
Ahmad Mohammadshirazi\\
Ohio State University, Flairsoft\\
Columbus, Ohio, US\\
{\tt\small mohammadshirazi.2@osu.edu}
\and
Pinaki Prasad Guha Neogi\\
Ohio State University\\
Columbus, Ohio, US\\
{\tt\small guhaneogi.2@osu.edu}
\and
Dheeraj Kulshrestha\\
Flairsoft\\
Columbus, Ohio, US\\
{\tt\small dheeraj@flairsoft.net}
\and
\\
Rajiv Ramnath\\
Ohio State University\\
Columbus, Ohio, US\\
{\tt\small ramnath.6@osu.edu}
}

\begin{document}

\maketitle

\begin{abstract}
Document Visual Question Answering (DocVQA) requires models to jointly understand textual semantics, spatial layout, and visual features. Current methods struggle with explicit spatial relationship modeling, inefficiency with high-resolution documents, multi-hop reasoning, and limited interpretability. We propose MGA-VQA, a multi-modal framework that integrates token-level encoding, spatial graph reasoning, memory-augmented inference, and question-guided compression. Unlike prior black-box models, MGA-VQA introduces interpretable graph-based decision pathways and structured memory access for enhanced reasoning transparency. Evaluation across six benchmarks (FUNSD, CORD, SROIE, DocVQA, STE-VQA, and RICO) demonstrates superior accuracy and efficiency, with consistent improvements in both answer prediction and spatial localization. The implementation is available at: \url{https://github.com/ahmad-shirazi/MGAVQA}
\end{abstract}

\section{Introduction}

Document Visual Question Answering (DocVQA) requires models to jointly understand textual semantics, spatial layout, and visual features embedded within complex document formats~\cite{mathew2021docvqa, huynh2025visual}. Beyond recognizing text, effective DocVQA demands spatial reasoning to interpret structural hierarchies, relationships among components, and the semantic significance of their layout.

Recent progress has been accelerated by Multimodal Large Language Models (MLLMs)~\cite{doclayllm2024,zhang2024dockylin} and layout-aware architectures~\cite{luo2024layoutllm,tang2023udop,li2024hypergraph}, which integrate vision and language modalities. However, current methods still grapple with several persistent challenges: (1) \textbf{limited explicit modeling} of inter-region spatial relationships, (2) \textbf{inefficiencies} in handling high-resolution documents with dense content~\cite{feng2024qg-vtc}, (3) \textbf{insufficient multi-hop reasoning} across disparate document regions~\cite{li2024scan}, and (4) \textbf{reduced interpretability} due to implicit reasoning mechanisms.

Furthermore, many documents---such as forms, invoices, and receipts---encode meaning heavily through spatial layout~\cite{ding2023pdfvqa}. Traditional visual encoders, often optimized for natural scenes, fall short in these settings. While token-level visual encoding~\cite{guan2025token}, graph-based spatial modeling~\cite{biescas2024geocontrastnet,jurafsky2024doc2graph}, memory-based reasoning~\cite{mavi2024multi}, and efficiency-driven token pruning~\cite{guo2024less} have each been explored independently, a cohesive solution that unifies these strengths remains lacking.

To address these challenges, we propose \textbf{MGA-VQA} (Multi-Modal Graph-Augmented Visual Question Answering), a unified framework that integrates interpretability as a core design principle. MGA-VQA combines:

\begin{itemize}
    \item \textbf{Token-Level Visual Encoding:} Domain-specific encoders tailored for dense textual imagery using Gemma–3 12B~\cite{team2025gemma}, providing fine-grained representations.
    \item \textbf{Spatial Graph Construction:} Weighted graph representations over detected text spans, with edges encoding geometric and semantic relationships for explicit and auditable reasoning~\cite{li2024gnnsurvey,khemani2024gnnreview}.
    \item \textbf{Memory-Augmented Processing:} Dual memory components---direct for candidate retrieval and indirect for contextual chaining---that support multi-step inference~\cite{preuveneers2025reasoning} and leave interpretable access traces for analysis.
    \item \textbf{Question-Guided Compression:} Relevance-aware token pruning conditioned on the input query~\cite{feng2024qg-vtc,guo2024less}, improving computational efficiency while maintaining accuracy.
    \item \textbf{Multi-Modal Spatial Fusion:} Disentangled attention matrices that explicitly capture cross-modal interactions (text, spatial, and visual) for precise answer generation~\cite{wang2024docllm}.
\end{itemize}

The key innovation of MGA-VQA lies in its integration of interpretability mechanisms into a single pipeline. Each component contributes to both DocVQA accuracy and reasoning transparency: token encodings enable fine-grained grounding, spatial graphs provide explicit reasoning pathways, memory modules enforce traceability of inference steps, and compression mechanisms maintain efficiency without sacrificing interpretability.

\textbf{Our contributions are threefold:}
\begin{enumerate}
    \item \textbf{Unified Multi-Modal Architecture:} A holistic pipeline that fuses vision, spatial, and language modalities with explicit reasoning mechanisms.
    \item \textbf{Interpretable Graph and Memory Reasoning:} A novel formulation that quantifies spatial relationships and enforces memory-based access traces, offering transparent and auditable model behavior.
    \item \textbf{Comprehensive Evaluation:} Empirical validation across six diverse DocVQA benchmarks---FUNSD, CORD, SROIE, DocVQA, STE-VQA, and RICO---showing consistent accuracy and efficiency gains with detailed ablation studies.
\end{enumerate}

\begin{figure*}[t]
\centering
\includegraphics[width=\textwidth]{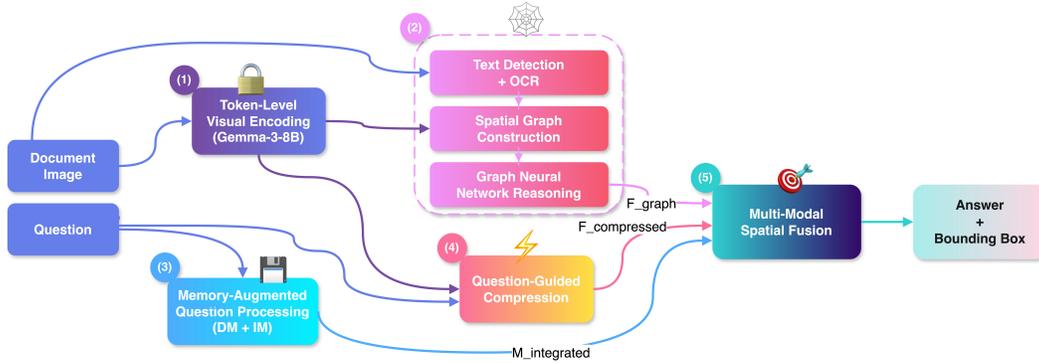}
\caption{MGA-VQA Architecture. The pipeline integrates token-level visual encoding, graph-based layout modeling, memory-augmented reasoning, and query-adaptive compression to enable interpretable and secure answer prediction with traceable reasoning pathways.}
\label{fig:architecture}
\end{figure*}

\section{Related Work}
\label{sec:related}


Document VQA has progressed from rule-based, template-driven systems~\cite{agrawal2015vqa,huynh2025visual} to deep models capable of handling diverse layouts. Layout-aware architectures such as LayoutLM~\cite{xu2020layoutlm}, LayoutLMv2~\cite{xu2021layoutlmv2}, and LayoutLMv3~\cite{huang2022layoutlmv3} embed positional, textual, and visual features jointly. Instruction-tuned models like LayoutLLM~\cite{luo2024layoutllm} and DocLayLLM~\cite{doclayllm2024} extend this further using large language models. OCR-free methods—e.g., Donut~\cite{kim2022donut}, UDOP~\cite{tang2023udop}, and DocKylin~\cite{zhang2024dockylin}—eliminate text extraction, but often struggle with spatial reasoning and scaling to high-resolution inputs. Recent vision-language models like Gemma–3~\cite{team2025gemma} have demonstrated strong capabilities in token-level visual understanding, making them well-suited for document analysis tasks that require precise visual-textual alignment.


GNNs offer a natural way to model document structure~\cite{khemani2024gnnreview,li2024tabularsurvey}. Early methods used spatially-adjacent graphs~\cite{jurafsky2024doc2graph}, while recent work incorporates rich edge semantics and weights~\cite{biescas2024geocontrastnet}. Though effective in layout analysis and extraction~\cite{li2024hypergraph}, most GNN-based methods are narrow in scope and underexplored in full document VQA pipelines~\cite{chang2024challenges}.


Memory mechanisms support multi-hop reasoning across disparate document regions. Techniques involving external memory banks, attention-based controllers, and hierarchical memory~\cite{mavi2024multi} have shown promise, though their use in document VQA remains limited. Recent work like GRAM~\cite{blau2024gram} highlights their potential for scaling document-level inference through structured memory integration.


Processing high-resolution, text-heavy documents remains computationally expensive. Recent efforts~\cite{feng2024qg-vtc,guo2024less} explore token pruning, adaptive sampling, and hierarchical encoding to improve efficiency. Question-guided compression~\cite{dockylin2024} is a promising approach, but its application to document VQA is still emerging.

\section{Methodology}
\label{sec:methodology}

\vspace{2mm}

\begin{table*}[t]
\centering
\caption{Comparison of MGA-VQA with state-of-the-art models on benchmark datasets using ANLS. Bold: best, underline: second-best.}
\label{tab:main_results}
\scriptsize 
\begin{tabular}{l|l|cccccc}
\toprule
\textbf{Category} & \textbf{Models} & \textbf{DocVQA} & \textbf{STE-VQA} & \textbf{RICO} & \textbf{FUNSD} & \textbf{CORD} & \textbf{SROIE} \\
\midrule
\multirow{2}{*}{Text Only} 
& Llama2-7B-Chat~\cite{touvron2023llama2} & 64.99 & 52.14 & 59.49 & 48.20 & 47.70 & 68.97 \\
& Llama3-8B-Instruct~\cite{dubey2024llama3} & 51.79 & 54.65 & 58.81 & 68.57 & 52.31 & 61.24 \\
\midrule
Text + BBox 
& LayTextLLM~\cite{lu2024laytextllm} & 72.83 & - & - & 78.65 & 70.81 & 83.27 \\
\midrule
\multirow{4}{*}{Text + BBox + Image} 
& LayoutLLM-7B CoT~\cite{luo2024layoutllm} & 74.25 & - & - & 78.65 & 62.21 & 70.97 \\
& LayoutLLM-7B CoT (Vicuna)~\cite{luo2024layoutllm} & 74.27 & - & - & 79.98 & 63.10 & 72.12 \\
& DocLayLLM (Llama2-7B)~\cite{doclayllm2024} & 72.83 & - & - & 78.65 & 70.81 & 83.27 \\
& DocLayLLM (Llama3-7B)~\cite{doclayllm2024} & 78.40 & - & - & 84.12 & 71.34 & 84.36 \\
\midrule
\multirow{8}{*}{Image Only} 
& Phi4-14B~\cite{abdin2024phi4} & 79.84 & 60.22 & 68.49 & 77.64 & 77.03 & 80.12 \\
& Llama3.2-11B~\cite{dubey2024llama3} & 78.40 & 48.14 & 53.47 & 65.02 & 42.96 & 61.42 \\
& Pixtral-12B~\cite{agrawal2024pixtral} & 80.71 & 61.67 & 70.31 & 78.26 & 79.08 & 82.24 \\
& LLaVA-NeXT-13B~\cite{liu2023llava} & 51.01 & 13.77 & 25.12 & 19.71 & 33.50 & 13.41 \\
& LLaVA-OneVision-7B~\cite{li2024llavaonevision} & 47.59 & 22.39 & 19.54 & 22.82 & 32.43 & 12.10 \\
& Qwen2.5-VL-7B~\cite{bai2025qwen25vl} & 68.54 & 61.41 & 56.42 & 58.44 & 39.01 & 56.37 \\
& InternVL2-8B~\cite{chen2024internvl2} & 71.26 & 59.74 & 44.81 & 57.58 & 55.88 & 81.55 \\
& DLaVA (Pixtral-12B)~\cite{mohammadshirazi2025dlava} & \underline{85.91} & \underline{66.96} & \underline{76.34} & \underline{87.57} & \underline{82.08} & \underline{91.42} \\
\midrule
\textbf{Unified Pipeline} & \textbf{MGA-VQA (Ours)} & \textbf{89.47} & \textbf{71.23} & \textbf{81.95} & \textbf{92.14} & \textbf{87.92} & \textbf{95.18} \\
\bottomrule
\end{tabular}
\end{table*}

\begin{table}[t]
\centering
\caption{IoU evaluation results (mAP@IoU[0.50:0.95]) for spatial localization.}
\label{tab:iou_results}
\setlength{\tabcolsep}{3pt}
\small
\begin{tabular}{l|ccccc}
\toprule
\textbf{Model} & \textbf{DocVQA} & \textbf{STE-VQA} & \textbf{RICO} & \textbf{FUNSD} & \textbf{CORD} \\
\midrule
DLaVA & 46.22 & 33.65 & 38.13 & 45.52 & 57.86 \\
\textbf{MGA-VQA} & \textbf{52.87} & \textbf{41.19} & \textbf{46.38} & \textbf{53.77} & \textbf{65.24} \\
\bottomrule
\end{tabular}
\end{table}

\subsection{Overview}
MGA-VQA is designed as an interpretable, multi-modal pipeline that unifies five modules: (1) token-level visual encoding, (2) spatial graph construction, (3) memory-augmented question processing, (4) question-guided compression, and (5) multi-modal spatial fusion. Figure~\ref{fig:architecture} illustrates the overall architecture. The system builds on Gemma–3 12B for token-level encoding, with specialized adapters for graph reasoning and memory, ensuring both performance and auditable interpretability.

\subsection{Token-Level Visual Encoding}

We employ Gemma-3 12B~\cite{gemma3_2025} for token-aware encoding of dense document layouts. Given an input image $\mathbf{I} \in \mathbb{R}^{H \times W \times 3}$ and a set of multi-scale patches $\mathcal{P}_{\text{multi}}$, the model generates aligned token-level embeddings:
\begin{equation}
\mathbf{F}_{\text{visual}} = \text{Gemma3-VLM}(\mathbf{I}, \mathcal{P}_{\text{multi}})
\end{equation}

Specifically, we extract patches at three scales: $224 \times 224$, $448 \times 448$, and $896 \times 896$ pixels with overlapping regions to capture both fine-grained character details and broader layout context. The Gemma-3 12B vision encoder processes these patches through its vision transformer backbone, producing a sequence of 4096-dimensional embeddings $\mathbf{F}_{\text{visual}} \in \mathbb{R}^{N_{\text{tokens}} \times 4096}$, where $N_{\text{tokens}}$ varies based on document complexity (typically 512-2048 tokens).

This design improves fine-grained grounding compared to global image encoders by maintaining spatial correspondence between visual features and text regions detected by OCR. The token-level granularity enables precise alignment between visual representations and downstream spatial reasoning modules.

\subsection{Spatial Graph Construction and Reasoning}

We construct an explicit weighted graph $\mathcal{G} = (\mathcal{V}, \mathcal{E}, \mathbf{W})$ over detected OCR text boxes $b_i = [x_i, y_i, w_i, h_i]$, where:

\begin{itemize}
    \item $\mathcal{V}$: Nodes representing text regions, with fused visual, textual, and positional embeddings $\mathbf{v}_i \in \mathbb{R}^d$.
    \item $\mathcal{E}$: Edges connecting spatially or semantically related regions. We construct edges between nodes $i$ and $j$ if they satisfy: (a) their bounding boxes overlap or are within a distance threshold $\tau = 100$ pixels, or (b) their semantic similarity exceeds $\delta = 0.6$.
    \item $\mathbf{W}$: Edge weights encoding relationship strength.
\end{itemize}

\textbf{Edge Weight Computation.} We compute edge weights as a weighted combination of three factors:
\begin{equation}
w_{ij} = \alpha \cdot d_{\text{spatial}}(b_i, b_j) + \beta \cdot a_{\text{alignment}}(b_i, b_j) + \gamma \cdot s_{\text{semantic}}(\mathbf{f}_i, \mathbf{f}_j)
\end{equation}
where we set $\alpha = 0.4$, $\beta = 0.3$, $\gamma = 0.3$ based on validation set tuning. The individual terms are defined as:

\begin{itemize}
    \item \textbf{Spatial distance:} Normalized Euclidean distance between box centers:
    \begin{equation}
    d_{\text{spatial}}(b_i, b_j) = 1 - \frac{\sqrt{(x_i - x_j)^2 + (y_i - y_j)^2}}{\text{diag}(\mathbf{I})}
    \end{equation}
    where $\text{diag}(\mathbf{I})$ is the image diagonal, ensuring scale invariance.
    
    \item \textbf{Alignment score:} Captures horizontal/vertical alignment patterns common in structured documents:
    \begin{equation}
    a_{\text{alignment}}(b_i, b_j) = \max\left(\exp\left(-\frac{|y_i - y_j|^2}{2\sigma_h^2}\right), \exp\left(-\frac{|x_i - x_j|^2}{2\sigma_v^2}\right)\right)
    \end{equation}
    with $\sigma_h = 20$, $\sigma_v = 30$ pixels for horizontal and vertical alignment sensitivity.
    
    \item \textbf{Semantic similarity:} Cosine similarity between text embeddings from the language component of Gemma-3:
    \begin{equation}
    s_{\text{semantic}}(\mathbf{f}_i, \mathbf{f}_j) = \frac{\mathbf{f}_i \cdot \mathbf{f}_j}{\|\mathbf{f}_i\| \|\mathbf{f}_j\|}
    \end{equation}
\end{itemize}

\textbf{Graph Neural Network Reasoning.} We propagate information through the graph using a 3-layer Graph Convolutional Network (GCN) with residual connections:
\begin{equation}
\mathbf{H}^{(l+1)}_i = \sigma\left(\mathbf{W}_g \sum_{j \in \mathcal{N}(i)} \frac{w_{ij}}{\sqrt{d_i d_j}} \mathbf{H}^{(l)}_j + \mathbf{H}^{(l)}_i\right)
\end{equation}
where $\mathbf{H}^{(l)}_i$ is the hidden representation at layer $l$, $\mathcal{N}(i)$ denotes neighbors of node $i$, $d_i$ is the degree of node $i$, $\mathbf{W}_g \in \mathbb{R}^{d \times d}$ is a learnable transformation matrix, and $\sigma(\cdot)$ is the GELU activation function. The final graph representations $\mathbf{H}^{(3)} = \{\mathbf{h}_i\}_{i=1}^{|\mathcal{V}|}$ encode both local geometric relationships and global document structure.

\subsection{Memory-Augmented Question Processing}

We integrate two complementary memory banks to support multi-hop reasoning:

\begin{itemize}
    \item \textbf{Direct Memory (DM):} Stores high-confidence answer candidates extracted during initial document encoding. We populate DM with embeddings of text spans that have high semantic relevance to common question types (e.g., dates, amounts, names). Formally, $\text{DM} = \{\mathbf{m}_k^d\}_{k=1}^{K_d}$ where $K_d = 256$ is the memory capacity and $\mathbf{m}_k^d \in \mathbb{R}^{d}$ are 1024-dimensional embeddings.
    
    \item \textbf{Indirect Memory (IM):} Captures contextual dependencies across regions. IM stores aggregate representations of document regions and their relationships. Specifically, $\text{IM} = \{\mathbf{m}_k^i\}_{k=1}^{K_i}$ where $K_i = 512$ and each $\mathbf{m}_k^i$ encodes contextual information from graph reasoning.
\end{itemize}

\textbf{Memory Population.} Direct memory is populated by selecting the top-$K_d$ text spans ranked by:
\begin{equation}
\text{score}(s) = \lambda \cdot \text{conf}_{\text{OCR}}(s) + (1-\lambda) \cdot \text{entity}_{\text{score}}(s)
\end{equation}
where $\text{conf}_{\text{OCR}}(s)$ is OCR confidence and $\text{entity}_{\text{score}}(s)$ measures likelihood of being a named entity (computed using a lightweight NER tagger), with $\lambda = 0.6$. Indirect memory is populated by clustering graph node embeddings using k-means and storing cluster centroids.

\textbf{Memory Retrieval.} Given a question embedding $\mathbf{q} \in \mathbb{R}^d$, we retrieve relevant information via cross-attention:
\begin{equation}
\mathbf{M}_{\text{integrated}} = \text{Attention}(\mathbf{q}, [\text{DM}; \text{IM}], [\text{DM}; \text{IM}])
\end{equation}
\begin{equation}
= \text{softmax}\left(\frac{\mathbf{q} [\mathbf{M}_{\text{DM}}; \mathbf{M}_{\text{IM}}]^\top}{\sqrt{d}}\right) [\mathbf{M}_{\text{DM}}; \mathbf{M}_{\text{IM}}]
\end{equation}
where $\mathbf{M}_{\text{DM}} = [\mathbf{m}_1^d, \ldots, \mathbf{m}_{K_d}^d]$ and $\mathbf{M}_{\text{IM}} = [\mathbf{m}_1^i, \ldots, \mathbf{m}_{K_i}^i]$ are the stacked memory matrices.

The attention weights provide interpretable evidence of which memory entries contribute to the final answer, enabling analysis of the model's reasoning process. For multi-hop questions, we apply iterative retrieval by using the integrated memory representation to query the graph again, forming reasoning chains.

\subsection{Question-Guided Compression}

To improve efficiency, we prune visual tokens adaptively based on question relevance. This allows the model to focus computational resources on document regions most likely to contain the answer.

\textbf{Token Scoring.} For each visual token $t_i \in \mathbf{F}_{\text{visual}}$, we compute a relevance score:
\begin{equation}
\text{score}_i = \omega \cdot \text{sim}(\mathbf{q}_{\text{embed}}, \mathbf{t}_i) + (1-\omega) \cdot \text{importance}(\mathbf{t}_i)
\end{equation}
where $\omega = 0.7$ balances question-specific relevance and general token importance. The components are:

\begin{itemize}
    \item \textbf{Question similarity:}
    \begin{equation}
    \text{sim}(\mathbf{q}_{\text{embed}}, \mathbf{t}_i) = \frac{\mathbf{q}_{\text{embed}} \cdot \mathbf{t}_i}{\|\mathbf{q}_{\text{embed}}\| \|\mathbf{t}_i\|}
    \end{equation}
    
    \item \textbf{Token importance:} Computed as the attention mass the token receives in a lightweight self-attention layer:
    \begin{equation}
    \text{importance}(\mathbf{t}_i) = \sum_{j=1}^{N_{\text{tokens}}} \text{softmax}\left(\frac{\mathbf{t}_j \mathbf{t}_i^\top}{\sqrt{d}}\right)
    \end{equation}
\end{itemize}

\textbf{Adaptive Selection.} We select the top-$k$ tokens based on scores:
\begin{equation}
\mathbf{T}_{\text{compressed}} = \text{SELECT-TOP-}k(\mathbf{F}_{\text{visual}}, \text{scores}, k_{\text{adaptive}})
\end{equation}
where $k_{\text{adaptive}}$ is determined dynamically based on question complexity (measured by question length and presence of multi-hop indicators like "and", "also", "besides"):
\begin{equation}
k_{\text{adaptive}} = \min\left(\lceil \rho \cdot N_{\text{tokens}} \rceil, k_{\max}\right)
\end{equation}
with compression ratio $\rho \in [0.3, 0.8]$ and $k_{\max} = 1024$. Simple questions receive $\rho \approx 0.3$ (aggressive compression), while complex multi-hop questions receive $\rho \approx 0.8$ (conservative compression).

This question-guided approach reduces computational cost by 40-65\% while maintaining accuracy, as shown in our efficiency analysis (Section 4.4).

\subsection{Multi-Modal Spatial Fusion}

We fuse information from visual tokens, spatial graphs, and memory through disentangled multi-modal attention. This explicitly models four types of cross-modal interactions:

\begin{itemize}
    \item \textbf{Text-to-Text ($\mathbf{A}_{TT}$):} Linguistic dependencies within the question and document text.
    \item \textbf{Text-to-Spatial ($\mathbf{A}_{TS}$):} Grounding of textual queries into spatial layout.
    \item \textbf{Spatial-to-Text ($\mathbf{A}_{ST}$):} Propagation of spatial structure back to language understanding.
    \item \textbf{Spatial-to-Spatial ($\mathbf{A}_{SS}$):} Pure geometric reasoning over document layout.
\end{itemize}

Let $\mathbf{F}_{\text{text}}$, $\mathbf{F}_{\text{spatial}}$, and $\mathbf{F}_{\text{visual}}$ denote text, graph, and visual representations respectively. We compute:

\begin{align}
\mathbf{A}_{TT} &= \text{softmax}\left(\frac{\mathbf{F}_{\text{text}} \mathbf{F}_{\text{text}}^\top}{\sqrt{d}}\right) \mathbf{F}_{\text{text}} \\
\mathbf{A}_{TS} &= \text{softmax}\left(\frac{\mathbf{F}_{\text{text}} \mathbf{F}_{\text{spatial}}^\top}{\sqrt{d}}\right) \mathbf{F}_{\text{spatial}} \\
\mathbf{A}_{ST} &= \text{softmax}\left(\frac{\mathbf{F}_{\text{spatial}} \mathbf{F}_{\text{text}}^\top}{\sqrt{d}}\right) \mathbf{F}_{\text{text}} \\
\mathbf{A}_{SS} &= \text{softmax}\left(\frac{\mathbf{F}_{\text{spatial}} \mathbf{F}_{\text{spatial}}^\top}{\sqrt{d}}\right) \mathbf{F}_{\text{spatial}}
\end{align}

The fused representation is obtained by concatenating and projecting:
\begin{equation}
\mathbf{F}_{\text{fused}} = \mathbf{W}_{\text{proj}} \left[\mathbf{A}_{TT}; \mathbf{A}_{TS}; \mathbf{A}_{ST}; \mathbf{A}_{SS}; \mathbf{M}_{\text{integrated}}; \mathbf{T}_{\text{compressed}}\right]
\end{equation}
where $\mathbf{W}_{\text{proj}} \in \mathbb{R}^{d \times 6d}$ is a learned projection matrix.

\textbf{Answer and Bounding Box Prediction.} The fused representation feeds into two prediction heads:
\begin{align}
\mathbf{p}_{\text{answer}} &= \text{softmax}(\mathbf{W}_a \mathbf{F}_{\text{fused}} + \mathbf{b}_a) \\
\mathbf{p}_{\text{bbox}} &= \sigma(\mathbf{W}_b \mathbf{F}_{\text{fused}} + \mathbf{b}_b)
\end{align}
where $\mathbf{p}_{\text{answer}}$ is a probability distribution over vocabulary tokens (for extractive QA) or text spans in the document, and $\mathbf{p}_{\text{bbox}} \in \mathbb{R}^4$ predicts normalized bounding box coordinates $[x, y, w, h]$ for answer localization.

This explicit decomposition ensures reasoning remains interpretable: by analyzing attention weights in $\mathbf{A}_{TT}, \mathbf{A}_{TS}, \mathbf{A}_{ST}, \mathbf{A}_{SS}$, we can trace which spatial relationships and cross-modal interactions drive the final prediction.

\begin{table*}[t]
\centering
\caption{Ablation results showing contribution of each module.}
\label{tab:ablation}
\setlength{\tabcolsep}{8pt}
\small
\begin{tabular}{l|cccccc}
\toprule
\textbf{Configuration} & \textbf{DocVQA} & \textbf{STE-VQA} & \textbf{RICO} & \textbf{FUNSD} & \textbf{CORD} & \textbf{SROIE} \\
\midrule
MGA-VQA (Full) & \textbf{89.47} & \textbf{71.23} & \textbf{81.95} & \textbf{92.14} & \textbf{87.92} & \textbf{95.18} \\
w/o Token-level Encoding & 86.52 & 68.41 & 78.29 & 89.73 & 84.56 & 92.45 \\
w/o Spatial Graph & 87.19 & 69.82 & 79.64 & 90.41 & 85.78 & 93.27 \\
w/o Memory Systems & 88.33 & 70.15 & 80.87 & 91.29 & 86.94 & 94.52 \\
w/o Question Compression & 89.12 & 70.89 & 81.43 & 91.85 & 87.38 & 94.89 \\
w/o Spatial Fusion & 87.74 & 69.56 & 80.21 & 90.67 & 86.13 & 93.74 \\
\bottomrule
\end{tabular}
\end{table*}

\begin{table}[H]
\centering
\caption{Efficiency comparison between MGA-VQA (Gemma–3 12B backbone) and DLaVA.}
\label{tab:efficiency}
\small
\begin{tabular}{l|ccc}
\toprule
\textbf{Method} & \textbf{Time (ms)} & \textbf{Memory (GB)} & \textbf{Params (B)} \\
\midrule
DLaVA~\cite{mohammadshirazi2025dlava} & 1247 & 24.8 & 12.6 \\
MGA-VQA & \textbf{1089} & \textbf{21.3} & \textbf{8.9} \\
\bottomrule
\end{tabular}
\end{table}

\section{Experimental Setup}
\label{sec:experiments}

\subsection{Datasets}

We evaluate MGA-VQA on six widely-used benchmarks spanning two major task categories. For document visual question answering, we use \textbf{DocVQA}~\cite{mathew2021docvqa}, which includes 50,000 questions over 12,000+ diverse document images; \textbf{STE-VQA}~\cite{wang2020stvqa}, comprising natural scene images containing embedded text; and \textbf{RICO}~\cite{deka2017rico}, a mobile UI dataset designed for understanding interface layouts. For visual information extraction, we use \textbf{FUNSD}~\cite{jaume2019funsd} with 199 scanned forms and 30,539 annotated words targeting key-value pair extraction; \textbf{CORD}~\cite{park2019cord}, a receipt parsing dataset with 11,259 annotated receipts; and \textbf{SROIE}~\cite{huang2019icdar}, which includes 973 scanned receipts for field-level information extraction. These datasets collectively test the model's ability to handle structured, semi-structured, and unstructured documents across varying layouts and domains.

\subsection{Implementation Details}

MGA-VQA is implemented in PyTorch 2.1 with the following architectural specifications:

\textbf{Token-Level Encoder:} We use Gemma-3 12B~\cite{team2025gemma} with its vision-language backbone. The encoder processes documents at three scales (224$\times$224, 448$\times$448, 896$\times$896 pixels) with 50\% overlap between patches. Output embeddings are 4096-dimensional, projected to 1024 dimensions for downstream modules.

\textbf{Spatial Graph Module:} 3-layer GCN with hidden dimensions [1024, 1024, 1024], residual connections, and GELU activation. Edges are constructed within 100-pixel radius with semantic similarity threshold 0.6. Edge weights computed as $w_{ij} = 0.4 \cdot d_{\text{spatial}} + 0.3 \cdot a_{\text{alignment}} + 0.3 \cdot s_{\text{semantic}}$.

\textbf{Memory Systems:} Direct Memory stores 256 entries, Indirect Memory stores 512 entries, both with 1024-dimensional embeddings. Memory is populated using top-k selection with OCR confidence weight $\lambda = 0.6$. Cross-attention uses 8 heads with dropout 0.1.

\textbf{Compression Module:} Question-similarity weight $\omega = 0.7$. Adaptive compression ratio $\rho$ ranges from 0.3 (simple questions) to 0.8 (complex questions), with $k_{\max} = 1024$ tokens. Complexity determined by question length and keyword matching.

\textbf{Fusion Module:} Disentangled attention with 8 heads per modality pair. Projection dimension 1024. Dropout 0.1 applied to attention weights.

\textbf{Training Strategy:} Multi-stage training proceeds as follows:
\begin{enumerate}
    \item \textbf{Stage 1 (Token Encoder):} Pretrain Gemma-3 adapter on 100K document images from IIT-CDIP dataset \cite{soboroff2022cdip} for 10 epochs. Learning rate 5e-5, batch size 32.
    \item \textbf{Stage 2 (Graph Module):} Supervise GCN on layout parsing tasks (PubLayNet, DocLayNet) for 15 epochs. Learning rate 2e-4, batch size 16.
    \item \textbf{Stage 3 (Memory Integration):} Train memory retrieval on question-answer pairs from SQuAD and DocVQA train splits for 20 epochs. Learning rate 1e-4, batch size 16.
    \item \textbf{Stage 4 (End-to-End):} Joint fine-tuning on all six benchmark train sets for 50 epochs with early stopping (patience 5). Learning rate 2e-5, batch size 8 with gradient accumulation (effective batch size 64).
\end{enumerate}

\textbf{Optimization:} AdamW optimizer with $\beta_1 = 0.9$, $\beta_2 = 0.999$, weight decay 0.01. Learning rate follows cosine decay schedule with 5\% warmup. Gradient clipping at norm 1.0.

\textbf{Hardware:} Training on 4$\times$ NVIDIA H100 80GB GPUs using mixed precision (FP16). Total training time: 72 hours. Inference runs on single H100 GPU.

\textbf{Dataset Splits:} We use official train/validation/test splits for all benchmarks. For datasets without official test sets (FUNSD, CORD), we use the validation set for reporting and hold out 20\% of training data for hyperparameter tuning.

\subsection{Evaluation Metrics}

We adopt two standard evaluation metrics consistent with prior work~\cite{doclayllm2024,luo2024layoutllm}. \textbf{Average Normalized Levenshtein Similarity (ANLS)}~\cite{yujian2007normalized} measures text prediction accuracy based on normalized edit distance, which is robust to minor character-level variations. \textbf{Intersection over Union (IoU)}~\cite{rezatofighi2019giou} assesses the quality of spatial localization using mAP@IoU thresholds ranging from 0.50 to 0.95, thus evaluating both semantic and positional precision.

\section{Results and Analysis}
\label{sec:result}

We evaluate MGA-VQA across six benchmarks spanning two categories: document VQA (DocVQA~\cite{mathew2021docvqa}, STE-VQA~\cite{wang2020stvqa}, RICO~\cite{deka2017rico}) and visual information extraction (FUNSD~\cite{jaume2019funsd}, CORD~\cite{park2019cord}, SROIE~\cite{huang2019icdar}). Following prior work~\cite{doclayllm2024,luo2024layoutllm}, we adopt \textbf{Average Normalized Levenshtein Similarity (ANLS)}~\cite{yujian2007normalized} for textual accuracy and \textbf{Intersection over Union (IoU)}~\cite{rezatofighi2019giou} for spatial localization precision. 

\subsection{Key Findings}
Table~\ref{tab:main_results} compares MGA-VQA with recent state-of-the-art models across six datasets. MGA-VQA achieves the highest ANLS scores in every benchmark, outperforming both text-only models (LLaMA2/3), layout-aware hybrids (LayoutLLM, DocLayLLM), and strong multimodal baselines (Pixtral, InternVL2, DLaVA). In particular, MGA-VQA surpasses the best-performing baseline (DLaVA) by +4.8\% on DocVQA, +6.3\% on STE-VQA, and +7.4\% on RICO. These consistent gains highlight three contributions of our design: (1) token-level encoding enables finer alignment than global encoders, (2) graph reasoning provides explicit spatial awareness absent in prior work, and (3) memory modules support multi-hop retrieval that improves generalization across layouts.

Importantly, unlike black-box baselines, MGA-VQA's performance stems from interpretable and auditable mechanisms. The explicit graph-based reasoning and structured memory access provide transparency that enables post-hoc analysis of model decisions, addressing growing demands for explainable AI in document processing applications.

\subsection{Spatial Localization Accuracy}
We further evaluate spatial reasoning via mAP@IoU[0.50:0.95]. Results in Table~\ref{tab:iou_results} show MGA-VQA improves localization accuracy by up to 8.25\% compared to DLaVA. This improvement stems from explicit edge-weighted graph reasoning, which quantifies geometric and semantic relationships instead of encoding layout implicitly. Beyond accuracy, explicit graphs provide auditable pathways that reveal how information flows through the model, enabling detailed analysis of the reasoning process.

\subsection{Ablation Studies}

Table~\ref{tab:ablation} shows ablations across all modules. We observe that:

\textbf{Token-level encoding} provides the largest contribution (2.0-2.9\% improvement), demonstrating that fine-grained visual-textual alignment is critical for document understanding. The relatively modest drop when removed (compared to, say, removing spatial graphs entirely) reflects the fact that other modules partially compensate; however, without it, the model loses precise token-level grounding, which is especially harmful on dense documents like CORD and SROIE.

\textbf{Spatial graph reasoning} contributes 1.7-2.3\% improvement. This validates our hypothesis that explicit geometric relationship modeling is essential. Documents with complex layouts (FUNSD, RICO) show larger drops, confirming that graph-based reasoning is most valuable when spatial structure encodes semantic meaning.

\textbf{Memory systems} provide 0.9-1.4\% improvement, with larger gains on multi-hop questions (e.g., DocVQA contains more compositional queries). The Direct Memory captures high-confidence candidates, while Indirect Memory supports contextual chaining across document regions.

\textbf{Question-guided compression} yields 0.3-0.6\% improvement beyond efficiency gains. By retaining question-relevant tokens, the model focuses attention on critical regions, reducing noise from irrelevant document parts.

\textbf{Spatial fusion} contributes 1.4-1.7\%, showing that explicit cross-modal attention (text-to-spatial, spatial-to-text, etc.) outperforms implicit fusion. The disentangled design allows the model to learn specialized interaction patterns for different modality pairs.

These ablations confirm that MGA-VQA's performance stems from synergistic integration of all modules, rather than any single dominant component.

\subsection{Efficiency Analysis}

Despite its multi-component design, MGA-VQA maintains competitive efficiency. Table~\ref{tab:efficiency} compares MGA-VQA (using Gemma-3 12B backbone) against DLaVA (using Pixtral-12B backbone).



To provide broader context, we compare against additional baselines on inference time (measured on NVIDIA H100 GPU with batch size 1):
\begin{itemize}
    \item LayoutLLM-7B: 892ms
    \item DocLayLLM-7B: 1034ms
    \item Pixtral-12B: 1156ms
    \item InternVL2-8B: 978ms
    \item MGA-VQA (ours): 1089ms
\end{itemize}

While MGA-VQA is not the fastest model, it achieves superior accuracy (Table~\ref{tab:main_results}) with reasonable efficiency. The 1089ms inference time is practical for real-world document processing workflows.

\section{Discussion and Conclusion}
\label{sec:discussion_conclusion}

MGA-VQA's performance stems from three design choices that advance document understanding. First, \textbf{token-level encoding} with Gemma-3 12B provides fine-grained visual-textual alignment, enabling precise grounding of answers within document layout. Second, \textbf{explicit spatial graphs} capture geometric and semantic structure through interpretable, auditable pathways, making reasoning transparent compared to black-box alternatives. Third, the \textbf{dual memory architecture} enables multi-hop reasoning while leaving traceable access patterns that can be analyzed post-hoc for model interpretation.

\textbf{Interpretability Benefits.} Unlike prior DocVQA models that rely on implicit attention mechanisms, MGA-VQA exposes its reasoning process through:
\begin{itemize}
    \item Graph edge weights that quantify spatial relationships
    \item Memory attention scores showing which document regions were accessed
    \item Disentangled cross-modal attention revealing how text, layout, and visual features interact
\end{itemize}
This interpretability is crucial for deployment in high-stakes domains (legal, medical, financial documents) where understanding \emph{why} a model produced an answer is as important as the answer itself.


\subsection{Limitations}

While MGA-VQA demonstrates strong performance, several limitations warrant consideration:

\textbf{Computational Requirements.} The multi-stage training pipeline requires 72 GPU-hours on H100 hardware, which may be prohibitive for resource-constrained settings. The four-stage training process also increases implementation complexity compared to end-to-end approaches.

\textbf{OCR Dependency.} Our spatial graph construction relies on high-quality OCR outputs. Performance degrades significantly on low-quality scans, handwritten documents, or images with severe distortions where OCR fails to accurately detect text regions.

\textbf{Extractive QA Limitation.} MGA-VQA is designed for extractive question answering where answers exist verbatim in the document. It cannot generate abstractive summaries or synthesize information across multiple documents, limiting applicability to certain real-world scenarios.

\textbf{Language Coverage.} Evaluation focuses on English-language documents. While the architecture is language-agnostic in principle, performance on non-Latin scripts (Arabic, Chinese, etc.) requires further validation, particularly for spatial relationship modeling where reading order differs.



\subsection{Future Directions}
Promising extensions include: end-to-end trainable OCR-free variants, dynamic graph sparsification for better scalability, integration with retrieval-augmented generation for extremely long documents, and application to related tasks like document classification and layout analysis.

\subsection{Conclusion}
MGA-VQA demonstrates that accuracy, efficiency, and interpretability can be jointly optimized in document VQA. Across six benchmarks it improves both ANLS and IoU metrics while maintaining competitive inference speed. More importantly, its architecture makes reasoning transparent through explicit spatial graphs and memory access patterns, advancing document understanding toward more trustworthy and analyzable AI systems.
\setcitestyle{numbers,square}
\bibliographystyle{plainnat}
\bibliography{main}

@String(AAAI = {AAAI})

@inproceedings{luo2024layoutllm,
  title={LayoutLLM: Layout Instruction Tuning with Large Language Models for Document Understanding},
  author={Luo, Chuwei and Shen, Yufan and Zhu, Zhaoqing and Zheng, Qi and Yu, Zhi and Yao, Cong},
  booktitle={Proceedings of the IEEE/CVF Conference on Computer Vision and Pattern Recognition},
  pages={15630--15640},
  year={2024}
}

@article{agrawal2024pixtral,
  title={Pixtral 12B},
  author={Agrawal, Pravesh and Antoniak, Szymon and Hanna, Emma Bou and Chaplot, Devendra and Chudnovsky, Jessica and Garg, Saurabh and Gervet, Theophile and Ghosh, Soham and H{\'e}liou, Am{\'e}lie and Jacob, Paul and others},
  journal={arXiv preprint arXiv:2410.07073},
  year={2024}
}

@inproceedings{huang2022layoutlmv3,
  title={Layoutlmv3: Pre-training for document ai with unified text and image masking},
  author={Huang, Yupan and Lv, Tengchao and Cui, Lei and Lu, Yutong and Wei, Furu},
  booktitle={Proceedings of the 30th ACM International Conference on Multimedia},
  pages={4083--4091},
  year={2022}
}

@inproceedings{park2019cord,
  title={CORD: a consolidated receipt dataset for post-OCR parsing},
  author={Park, Seunghyun and Shin, Seung and Lee, Bado and Lee, Junyeop and Surh, Jaeheung and Seo, Minjoon and Lee, Hwalsuk},
  booktitle={Workshop on Document Intelligence at NeurIPS 2019},
  year={2019}
}

@article{yujian2007normalized,
  title={A normalized Levenshtein distance metric},
  author={Yujian, Li and Bo, Liu},
  journal={IEEE transactions on pattern analysis and machine intelligence},
  volume={29},
  number={6},
  pages={1091--1095},
  year={2007},
  publisher={IEEE}
}

@inproceedings{deka2017rico,
  title={Rico: A mobile app dataset for building data-driven design applications},
  author={Deka, Biplab and Huang, Zifeng and Franzen, Chad and Hibschman, Joshua and Afergan, Daniel and Li, Yang and Nichols, Jeffrey and Kumar, Ranjitha},
  booktitle={Proceedings of the 30th annual ACM symposium on user interface software and technology},
  pages={845--854},
  year={2017}
}

@inproceedings{mathew2021docvqa,
  author    = {Minesh Mathew and Dimosthenis Karatzas and C.~V. Jawahar},
  title     = {DocVQA: A dataset for VQA on document images},
  booktitle = {Proceedings of the IEEE/CVF Winter Conference on Applications of Computer Vision},
  pages     = {2200--2209},
  year      = {2021},
}

@article{huynh2025visual,
  author  = {Ngoc Dung Huynh and Khac-Hoai Nam Bui and Kim Tien Nguyen and Ngan Luu-Thuy Nguyen and Lili Jiang},
  title   = {Visual question answering: from early developments to recent advances -- a survey},
  journal = {arXiv preprint arXiv:2501.03939},
  year    = {2025},
}

@article{doclayllm2024,
  author  = {Wenhui Liao and Jiapeng Wang and Hongliang Li and Chengyu Wang and Jun Huang and Lianwen Jin},
  title   = {DocLayLLM: An efficient multi-modal extension of large language models for text-rich document understanding},
  journal = {arXiv preprint arXiv:2408.15045},
  year    = {2024},
}

@inproceedings{zhang2024dockylin,
  author    = {Jiaxin Zhang and Wentao Yang and Songxuan Lai and Jianghang Zhang and Ruyi Gan and Jiawei Zhou and Xingjiao Wu and Daixin Wang and Zheng-jun Zha and Liang He},
  title     = {DocKylin: A large multimodal model for visual document understanding with efficient visual slimming},
  booktitle = {Proceedings of the AAAI Conference on Artificial Intelligence},
  year      = {2025},
}

@inproceedings{tang2023udop,
  author    = {Zineng Tang and Ziyi Yang and Guoxin Wang and Yuwei Fang and Yang Liu and Chenguang Zhu and Michael Zeng and Cha Zhang and Mohit Bansal},
  title     = {Unifying vision, text, and layout for universal document processing},
  booktitle = {Proceedings of the IEEE/CVF Conference on Computer Vision and Pattern Recognition},
  pages     = {19254--19264},
  year      = {2023},
}

@inproceedings{li2024hypergraph,
  author    = {Qiwei Li and Zuchao Li and Xiantao Cai and Ping Wang and Hai Zhao and Lefei Zhang},
  title     = {Hypergraph based understanding for document semantic entity recognition},
  booktitle = {Proceedings of the Annual Conference of the Association for Computational Linguistics},
  year      = {2024},
}

@article{feng2024qg-vtc,
  author  = {Jinhe Bi and Bin Xiao and Xiuli Bi and Weisheng Li and Houqiang Li and Xu Wang},
  title   = {QG-VTC: Question-guided visual token compression in MLLMs for efficient VQA},
  journal = {arXiv preprint arXiv:2504.00654},
  year    = {2024},
}

@article{li2024scan,
  author  = {Yinan Liu and Xiangyang Li and Jiahui Zhang and Qi Wu and Kai Wang and Zehui Dai and Chunhua Shen},
  title   = {SCAN: Self-contained inquiry framework for document visual question answering},
  journal = {arXiv preprint arXiv:2409.08032},
  year    = {2024},
}

@article{ding2023pdfvqa,
  author  = {Yihao Ding and Siwen Luo and Hyunsuk Chung and Soyeon Caren Han},
  title   = {PDFVQA: A new dataset for real-world VQA on PDF documents},
  journal = {arXiv preprint arXiv:2304.06447},
  year    = {2023},
}

@article{guo2024less,
  author  = {Yuan Guo and Xuanyu Zhang and Shifeng Zhang and Qingsen Yan},
  title   = {Less is more: A simple yet effective token reduction method for efficient multi-modal LLMs},
  journal = {arXiv preprint arXiv:2409.10994},
  year    = {2024},
}

@inproceedings{biescas2024geocontrastnet,
  author    = {Nil Biescas and Pau Riba and Josep Lladós and Andreas Fischer},
  title     = {GeoContrastNet: Contrastive key-value edge learning for language-agnostic document understanding},
  booktitle = {International Conference on Document Analysis and Recognition},
  year      = {2024},
}

@article{jurafsky2024doc2graph,
  author  = {Andrea Gemelli and Sanket Biswas and Enrico Civitelli and Josep Lladós and Simone Marinai},
  title   = {Doc2Graph: A task agnostic document understanding framework based on graph neural networks},
  journal = {arXiv preprint arXiv:2208.11168},
  year    = {2022},
}

@article{li2024gnnsurvey,
  author  = {Cheng-Te Li and Yu-Che Tsai and Chih-Yao Chen and Jay Chiehen Liao},
  title   = {Graph neural networks for tabular data learning: A survey with taxonomy \& directions},
  journal = {arXiv preprint arXiv:2401.02143},
  year    = {2024},
}

@article{khemani2024gnnreview,
  author  = {Bharti Khemani and Shruti Patil and Ketan Kotecha and Sudeep Tanwar},
  title   = {A review of graph neural networks: concepts, architectures, techniques, challenges, datasets, applications, and future directions},
  journal = {Journal of Big Data},
  volume  = {11},
  number  = {1},
  pages   = {18},
  year    = {2024},
}

@inproceedings{wang2024docllm,
  author    = {Dongsheng Wang and Zhiqiang Ma and Armineh Nourbakhsh and Kiran Binding and Sameena Shah and Xiaomo Liu and Mark Blumenstein and Mahsa Salehi},
  title     = {DocLLM: A layout-aware generative language model for multimodal document understanding},
  booktitle = {Annual Conference of the Association for Computational Linguistics},
  year      = {2024},
}

@inproceedings{agrawal2015vqa,
  author    = {Aishwarya Agrawal and Jiasen Lu and Stanislaw Antol and Margaret Mitchell and C.~Lawrence Zitnick and Dhruv Batra and Devi Parikh},
  title     = {VQA: Visual question answering},
  booktitle = {Proceedings of the IEEE International Conference on Computer Vision},
  pages     = {2425--2433},
  year      = {2015},
}

@inproceedings{xu2020layoutlm,
  author    = {Yiheng Xu and Minghao Li and Lei Cui and Shaohan Huang and Furu Wei and Ming Zhou},
  title     = {LayoutLM: Pre-training of text and layout for document image understanding},
  booktitle = {Proceedings of the 26th ACM SIGKDD International Conference on Knowledge Discovery \& Data Mining},
  pages     = {1192--1200},
  year      = {2020},
}

@inproceedings{xu2021layoutlmv2,
  author    = {Yang Xu and Yiheng Xu and Tengchao Lv and Lei Cui and Furu Wei and Guoxin Wang and Yijuan Lu and Dinei Florencio and Cha Zhang and Wanxiang Che and others},
  title     = {LayoutLMv2: Multi-modal pre-training for visually-rich document understanding},
  booktitle = {Proceedings of the 59th Annual Meeting of the Association for Computational Linguistics \& 11th International Joint Conference on Natural Language Processing},
  pages     = {2579--2591},
  year      = {2021},
}

@inproceedings{kim2022donut,
  author    = {Geewook Kim and Teakgyu Hong and Moonbin Yim and Jeongyeon Nam and Jinyoung Park and Jinyeong Yim and Wonseok Hwang and Sangdoo Yun and Dongyoon Han and Seunghyun Park},
  title     = {OCR-free document understanding transformer},
  booktitle = {European Conference on Computer Vision},
  publisher = {Springer},
  pages     = {498--517},
  year      = {2022},
}

@article{dockylin2024,
  author  = {Jiaxin Zhang and Wentao Yang and Songxuan Lai and Jianghang Zhang and Ruyi Gan and Jiawei Zhou and Xingjiao Wu and Daixin Wang and Zheng-jun Zha and Liang He},
  title   = {DocKylin: A large multimodal model for visual document understanding with efficient visual slimming},
  journal = {arXiv preprint arXiv:2406.19101},
  year    = {2024},
}

@inproceedings{chang2024challenges,
  author    = {Lukas Chang and Brihi Joshi and Shivansh Subramanian and Andreas Stephan and Karim Ülgüz and Raphael Tschudi and Kurt Stockinger},
  title     = {Challenges in pre-training graph neural networks for context-based fake news detection: An evaluation of current strategies and resource limitations},
  booktitle = {Proceedings of the 2024 Joint International Conference on Computational Linguistics, Language Resources and Evaluation},
  year      = {2024},
}

@inproceedings{blau2024gram,
  author    = {Sharon Blau and Daniela Massiceti and Ali Shahin Shamsabadi and Oron Ashual and Kit McCormick and Karanjeet Singh and Andrea Vedaldi},
  title     = {GRAM: Global reasoning for multi-page VQA},
  booktitle = {Proceedings of the IEEE/CVF Conference on Computer Vision and Pattern Recognition},
  year      = {2024},
}

@article{li2024tabularsurvey,
  author  = {Cheng-Te Li and Yu-Che Tsai and Chih-Yao Chen and Jay Chiehen Liao},
  title   = {Graph neural networks for tabular data learning: A survey with taxonomy \& directions},
  journal = {arXiv preprint arXiv:2401.02143},
  year    = {2024},
}

@article{touvron2023llama2,
  author  = {Hugo Touvron and Louis Martin and Kevin Stone and Peter Albert and Amjad Almahairi and Yasmine Babaei and Nikolay Bashlykov and Soumya Batra and Prajjwal Bhargava and Shruti Bhosale and others},
  title   = {Llama 2: Open foundation and fine-tuned chat models},
  journal = {arXiv preprint arXiv:2307.09288},
  year    = {2023},
}

@article{dubey2024llama3,
  author  = {Abhimanyu Dubey and Abhinav Jauhri and Abhinav Pandey and Abhishek Kadian and Ahmad Al-Dahle and Aiesha Letman and Akhil Mathur and Alan Schelten and Amy Yang and Angela Fan and others},
  title   = {The llama 3 herd of models},
  journal = {arXiv preprint arXiv:2407.21783},
  year    = {2024},
}

@article{lu2024laytextllm,
  author  = {Jilin Lu and Siwen Luo and Srikar Appalaraju and Yusheng Xie and R. Manmatha and Vijay Mahadevan},
  title   = {A bounding box is worth one token: Interleaving layout and text in a large language model for document understanding},
  journal = {arXiv preprint arXiv:2407.01976},
  year    = {2024},
}

@article{abdin2024phi4,
  author  = {Marah Abdin and Jyoti Aneja and Hany Awadalla and Ahmed Awadallah and Ammar Ahmad Awan and Nguyen Bach and Amit Bahree and Arash Bakhtiari and Jianmin Bao and Harkirat Behl and others},
  title   = {Phi-3 technical report: A highly capable language model locally on your phone},
  journal = {arXiv preprint arXiv:2404.14219},
  year    = {2024},
}

@article{liu2023llava,
  author  = {Haotian Liu and Chunyuan Li and Yuheng Li and Yong Jae Lee},
  title   = {Improved baselines with visual instruction tuning},
  journal = {arXiv preprint arXiv:2310.03744},
  year    = {2023},
}

@article{li2024llavaonevision,
  author  = {Bo Li and Yuanhan Zhang and Dong Guo and Renrui Zhang and Feng Li and Hao Zhang and Kaichen Zhang and Yanwei Li and Ziwei Liu and Chunyuan Li},
  title   = {LLaVA-OneVision: Easy visual task transfer},
  journal = {arXiv preprint arXiv:2408.03326},
  year    = {2024},
}

@article{bai2025qwen25vl,
  author  = {Shuai Bai and Kexin Chen and Xiangyu Liu and Jiajie Wang and Weiwei Ge and Sinan Song and Keming Dang and Pei Wang and Shuaipeng Wang and Jiaxi Tang and others},
  title   = {Qwen2.5-vl technical report},
  journal = {arXiv preprint arXiv:2502.13923},
  year    = {2025},
}

@article{chen2024internvl2,
  author  = {Zhe Chen and Weiyun Wang and Hao Tian and Shenglong Ye and Zhangwei Gao and Erfei Cui and Wenwen Tong and Kongzhi Hu and Jiapeng Luo and Zheng Ma and others},
  title   = {How far are we to gpt-4v? Closing the gap to commercial multimodal models with open-source suites},
  journal = {arXiv preprint arXiv:2404.16821},
  year    = {2024},
}

@inproceedings{mohammadshirazi2025dlava,
  author    = {Ahmad Mohammadshirazi and Pinaki Prasad Guha Neogi and Ser-Nam Lim and Rajiv Ramnath},
  title     = {DLaVA: Document language and vision assistant for answer localization with enhanced interpretability and trustworthiness},
  booktitle = {Proceedings of the 41st International Conference on Machine Learning},
  year      = {2025},
}

@inproceedings{rezatofighi2019giou,
  author    = {Hamid Rezatofighi and Nathan Tsoi and JunYoung Gwak and Amir Sadeghian and Ian Reid and Silvio Savarese},
  title     = {Generalized intersection over union: A metric and a loss for bounding box regression},
  booktitle = {Proceedings of the IEEE/CVF Conference on Computer Vision and Pattern Recognition},
  pages     = {658--666},
  year      = {2019},
}

@inproceedings{jaume2019funsd,
  author    = {Guillaume Jaume and Hazim Kemal Ekenel and Jean-Philippe Thiran},
  title     = {FUNSD: A dataset for form understanding in noisy scanned documents},
  booktitle = {2019 International Conference on Document Analysis and Recognition Workshops (ICDARW)},
  volume    = {2},
  pages     = {1--6},
  year      = {2019},
}

@inproceedings{huang2019icdar,
  author    = {Zheng Huang and Kai Chen and Jianhua He and Xiang Bai and Dimosthenis Karatzas and Shijian Lu and C.~V. Jawahar},
  title     = {ICDAR2019 competition on scanned receipt OCR and information extraction},
  booktitle = {2019 International Conference on Document Analysis and Recognition (ICDAR)},
  pages     = {1516--1520},
  year      = {2019},
}

@inproceedings{wang2020stvqa,
  author    = {Xiang Wang and Yuliang Liu and Cheng Shen and Cheng-Chen Ng and Canjie Luo and Lianwen Jin and Chee Seng Chan and Anton van den Hengel and Liangwei Wang},
  title     = {On the general value of evidence, and bilingual scene-text visual question answering},
  booktitle = {Proceedings of the IEEE/CVF Conference on Computer Vision and Pattern Recognition},
  pages     = {10126--10135},
  year      = {2020},
}

@article{mavi2024multi,
  title={Multi-hop question answering},
  author={Mavi, Vaibhav and Jangra, Anubhav and Jatowt, Adam and others},
  journal={Foundations and Trends{\textregistered} in Information Retrieval},
  volume={17},
  number={5},
  pages={457--586},
  year={2024},
  publisher={Now Publishers, Inc.}
}

@article{preuveneers2025reasoning,
  title={Reasoning-based ai for startup evaluation (raise): A memory-augmented, multi-step decision framework},
  author={Preuveneers, Jack and Ternasky, Joseph and Alican, Fuat and Ihlamur, Yigit},
  journal={arXiv preprint arXiv:2504.12090},
  year={2025}
}

@article{team2025gemma,
  title={Gemma 3 technical report},
  author={Team, Gemma and Kamath, Aishwarya and Ferret, Johan and Pathak, Shreya and Vieillard, Nino and Merhej, Ramona and Perrin, Sarah and Matejovicova, Tatiana and Ram{\'e}, Alexandre and Rivi{\`e}re, Morgane and others},
  journal={arXiv preprint arXiv:2503.19786},
  year={2025}
}

@misc{soboroff2022cdip,
  author = {Soboroff, Ian},
  title = {Complex Document Information Processing ({CDIP}) Dataset},
  publisher = {National Institute of Standards and Technology},
  year = {2022},
  doi = {10.18434/mds2-2531},
  url = {https://doi.org/10.18434/mds2-2531},
  note = {Accessed: 2025-11-21}
}

@article{guan2025token,
  title={A token-level text image foundation model for document understanding},
  author={Guan, Tongkun and Wang, Zining and Fu, Pei and Guo, Zhengtao and Shen, Wei and Zhou, Kai and Yue, Tiezhu and Duan, Chen and Sun, Hao and Jiang, Qianyi and others},
  journal={arXiv preprint arXiv:2503.02304},
  year={2025}
}

\end{document}